\documentclass{article}

\usepackage{svg}
\usepackage{microtype}
\usepackage{graphicx}
\usepackage{bm}
\usepackage{amsmath}
\usepackage{amssymb}
\usepackage{amsthm}
\usepackage{amsfonts}
\usepackage{multirow}
\usepackage{xcolor}
\usepackage{colortbl} 
\usepackage{array}
\usepackage{booktabs}
\usepackage{boldline} 
\usepackage{hyperref}
\usepackage[toc,page]{appendix}

\newtheorem{theorem}{Theorem}[section]

\usepackage[accepted]{icml2020}

\icmltitlerunning{Deep Divergence Learning}

\begin{document}

\twocolumn[
\icmltitle{Deep Divergence Learning}


\begin{icmlauthorlist}
\icmlauthor{Kubra Cilingir}{to}
\icmlauthor{Rachel Manzelli}{to}
\icmlauthor{Brian Kulis}{to}
\end{icmlauthorlist}

\icmlaffiliation{to}{Department of Electrical and Computer Engineering, Boston University, Boston, Massachusetts, USA}

\icmlcorrespondingauthor{Kubra Cilingir}{kubra@bu.edu}
\icmlcorrespondingauthor{Rachel Manzelli}{manzelli@bu.edu}
\icmlcorrespondingauthor{Brian Kulis}{bkulis@bu.edu}

\icmlkeywords{Metric Learning, Bregman Divergences, Deep Learning, Triplet Loss, Contrastive Loss}

\vskip 0.3in
]

\printAffiliationsAndNotice{}

\begin{abstract}
Classical linear metric learning methods have recently been extended along two distinct lines: deep metric learning methods for learning embeddings of the data using neural networks,
and Bregman divergence learning approaches for extending learning Euclidean distances to more general divergence measures such as divergences over distributions.  In this paper, we introduce \textit{deep Bregman divergences}, which are based on learning and parameterizing functional Bregman divergences using neural networks, and which unify and extend these existing lines of work.  We show in particular how deep metric learning formulations, kernel metric learning, Mahalanobis metric learning, and moment-matching functions for comparing distributions arise as special cases of these divergences in the symmetric setting.  We then describe a deep learning framework for learning general functional Bregman divergences, and show in experiments that this method yields superior performance on benchmark datasets as compared to existing deep metric learning approaches.  We also discuss novel applications, including a semi-supervised distributional clustering problem, and a new loss function for unsupervised data generation.
\end{abstract}

\vspace{-.2cm}
\section{Introduction}
\vspace{-.2cm}
The goal of metric learning is to use supervised data in order to learn a distance function (or more general divergence measure) that is tuned to the data and task at hand.  Classical approaches to metric learning are generally focused on the linear regime, where one learns a linear mapping of the data and then applies the Euclidean distance in the mapped space for downstream tasks such as clustering, ranking, and classification~\cite{icml,weinberger_jmlr09,goldberger_nips04}.  These methods, known as Mahalanobis metric learning approaches, have been analyzed theoretically, are scalable, and usually involve convex optimization problems that can be solved globally~\cite{kulis2013metric,bellet2015metric}.

Classical metric learning methods have been extended along various axes; two important directions are deep metric learning and Bregman divergence learning.  Deep metric learning approaches replace the linear mapping learned in Mahalanobis metric learning methods with more general mappings that are learned via neural networks~\cite{hoffer2015deep,chopra_cvpr05}.  On the other hand, Bregman divergence methods replace the squared Euclidean distance with arbitrary Bregman divergences~\cite{bregman_67}, and learn the underlying generating function of the Bregman divergence via piecewise linear approximators~\cite{siahkamari_arxiv} or convex combinations of existing basis functions~\cite{wu2009learning}.  
These two extensions of classical metric learning are complementary and disjoint.  For instance, Bregman divergence approaches can be utilized in scenarios where one needs to compare distributions (the well-known KL-divergence arises as a special case), but the learning problems are not directly applicable to the deep learning setting.  Similarly, deep metric learning methods still employ Euclidean distances, and are thus not directly amenable to problems where one needs to compare distributions.

\begin{figure}[t]
\centering
\includegraphics[width=8.5cm]{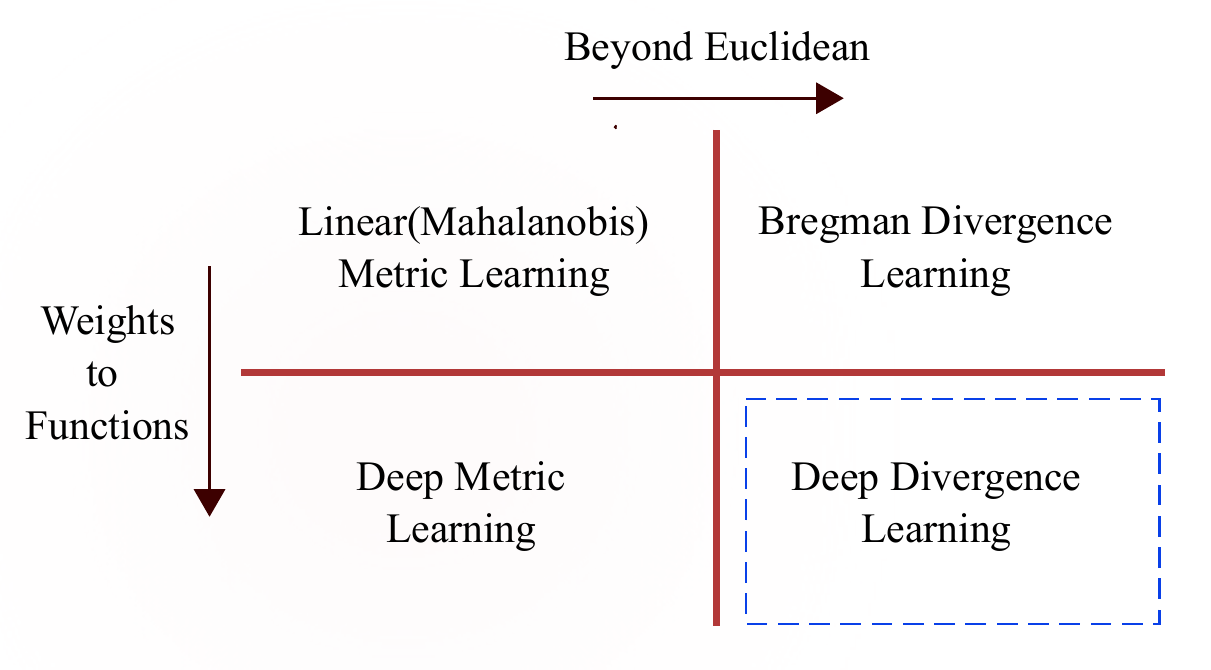}
\caption{Overview of our framework in comparison to existing metric learning approaches.  Deep Bregman divergences feature both the ability to learn divergences beyond Euclidean (such as divergences over distributions) while encompassing parameterizations that are amenable to deep learning architectures.}
\label{fig:overview}
\end{figure}

In this paper, we introduce a framework for studying Bregman divergences that can naturally be learned in the deep setting.  Figure~\ref{fig:overview} gives a high-level overview of our approach, which we term as \textit{deep Bregman divergences}, in comparison to existing metric learning approaches.  These divergences are based on functional Bregman divergences~\cite{functional_bd}, which were introduced as a extension of classical Bregman
divergences but with functional inputs instead of vector inputs.  In this functional setting, the underlying Bregman divergence is parameterized by a convex functional whose input itself is a function.  

We first perform an analysis for the symmetric divergence case.  In this setting, we prove a result about the form for any functional Bregman divergence and observe that many existing metric learning models can be seen to arise from special cases of this form.  These include deep learning methods, classical linear metric learning methods, and kernel metric learning.  There are also special cases that include moment-matching functions, which yields connections to the Wasserstein distance~\cite{arjovsky_arxiv17}, maximum mean discrepancy (MMD), and kernel MMD~\cite{Gretton2012}.

We then turn our attention to the strictly more general case, where the divergences need not be symmetric; the KL-divergence is a classical example of such an asymmetric Bregman divergence.  In this setting, we describe a framework for learning an arbitrary deep Bregman divergence.  Our approach is based on appropriately parameterizing the convex functional governing the underlying Bregman divergence with a neural network, and learning the resulting parameters of that network. 

We describe several applications of our proposed deep Bregman divergence framework.  First, we can extend existing deep metric learning formulations to learn more general deep Bregman divergences.  Second, since our divergences can naturally be applied to compare distributions on data, another application is in unsupervised generative learning, where the goal is to minimize a learned distributional divergence between real and generated data.  In particular, we discuss connections to GAN models and describe some novel algorithms for unsupervised data generation.  Third, we describe a semi-supervised distributional clustering problem.  Here, the problem is to cluster data where each data point is represented as a distribution---for example, a movie's rating may be represented as a distribution over user scores---using training data where we know whether pairs of distributions should be clustered together or not.

In all three of the above settings, we show empirical results that highlight the benefits of our framework.  In particular, we show that learning asymmetric divergences offers performance gains over existing symmetric models on benchmark data, and achieve state-of-the-art classification performance in some settings.  We also show that our clustering algorithm outperforms existing baselines on a simple proof-of-concept dataset as well as several human activity sensor data sets, and that our data generation results suggest that there may be value in further developing and studying new learned distributional divergence measures.

\vspace{-.2cm}
\section{Related Work}
\vspace{-.2cm}
Much of the early work on metric learning focused on the \textit{linear} setting, often referred to as Mahalanobis metric learning.  In this setting, the goal is to learn a global linear transformation of the data and apply standard distances such as the Euclidean distance on top of the learned transformation.  This is often expressed as learning a distance function of the form $d_A(\bm{x},\bm{y}) = (\bm{x}-\bm{y})^T A (\bm{x} - \bm{y})$, where $A$ is a positive semi-definite matrix.  This is equivalent to learning a linear transformation $G$, where $A = G^T G$, since $d_A(\bm{x},\bm{y}) = (\bm{x} - \bm{y})^T G^T G (\bm{x} - \bm{y}) = \|G \bm{x} - G \bm{y}\|_2^2$.  Examples of this approach to metric learning include MMC~\cite{xing2003distance}, MCML~\cite{globerson_nips05}, LMNN~\cite{weinberger_jmlr09}, ITML~\cite{icml}, POLA~\cite{shalev2004online}, LEGO~\cite{jain_nips08}, and others.  See the surveys by~\citet{kulis2013metric} and~\citet{bellet2015metric} for further references and details on some of these approaches.  Note that one of the advantages of the linear approach is that one can often provide performance guarantees---for instance, a significant amount of work has gone into proving regret bounds in the online setting~\cite{shalev2004online}, as well as generalization bounds~\cite{bellet2015robustness,cao2016generalization} for some Mahalanobis metric learning models.

While linear methods are simpler and can often be analyzed theoretically, in practice it is often useful to learn other, non-linear, approaches to metric learning. 
For instance, one can show that many linear models can be appropriately adapted to run in kernel space~\cite{kernel_metric,chatpatanasiri_neurocomputing10}.  
Another more recent approach to moving beyond linear metric learning is the Bregman divergence learning framework discussed in the introduction~\cite{siahkamari_arxiv,wu2009learning}.  Here we move beyond learning Mahalanobis metrics, but instead focus on a strictly larger class of divergences that includes asymmetric divergences such as the KL-divergence, Itakura-Saito divergence, and others.  These may be considered as non-linear approaches (since the resulting divergence does not involve linear transformations in general).  The Bregman learning framework is thus more powerful than linear approaches but also remains well-principled: one can prove generalization bounds in this framework.

The third, and by far the most well-studied, approach to non-linear metric learning is known as deep metric learning, and involves learning a neural network to embed data into some new space, where standard distances such as Euclidean distance are used.  If $f$ is a function that maps an input $\bm{x}$ to an embedding $f(\bm{x})$, then the resulting learned metric is typically $\|f(\bm{x}) - f(\bm{y})\|_2^2$.  Several popular loss functions have been proposed to learn such a metric---the two main ones are the contrastive loss~\cite{chopra_cvpr05} and the triplet loss~\cite{hoffer2015deep}.  Both utilize supervision (pairwise for the contrastive loss and relative constraints for the triplet loss) and use the learned distance $\|f(\bm{x}) - f(\bm{y})\|_2^2$.  Moreover, there has been considerable follow-up work that explores how best to choose pairs or triples of points from a training set to achieve the best results~\cite{HermansBeyer2017Arxiv}.  There has also been work on deep metric learning using other losses, such as the angular loss~\cite{angular_metric_learning} or the average precision for deep metric learning to rank~\cite{deep_ranking}. 

Our work also has ties to methods involving comparing distributions.  Examples of such measures that are relevant to our work include the maximum mean discrepancy metric (also known as the integral probability metric)~\cite{Gretton2012}, the kernel MMD, and the Wasserstein distance~\cite{arjovsky_arxiv17}.  Several notions of divergences over distributions have been used for unsupervised data generation in GAN-type models, including the Jensen-Shannon divergence~\cite{gans}, the Wasserstein distance~\cite{arjovsky_arxiv17}, and the MMD~\cite{gmnn,mmd_gan}.

\vspace{-.2cm}
\section{Deep Bregman Divergences}
\vspace{-.2cm}
We now turn our attention to functional Bregman divergences, the main tool for our learning problems.  Our goal is two-fold: we first prove a result that characterizes the form for a symmetric functional Bregman divergence and show connections between this form and existing metric learning models.  Second, we consider a parameterization for arbitrary functional Bregman divergences that will permit learning via neural networks.

\subsection{Bregman Divergences and Functional Bregman Divergences}
A Bregman divergence is a generalized measure of distance between objects, parameterized by a strictly convex function $\phi$~\cite{bregman_67}.  Let $\phi: \Omega \rightarrow \mathbb{R}$, where $\Omega$ is a closed, convex set.  The Bregman divergence with respect to $\phi$ (for vector inputs) is defined as
\begin{displaymath}
D_{\phi}(\bm{x}, \bm{y}) = \phi(\bm{x}) - \phi(\bm{y}) - (\bm{x} - \bm{y})^T \nabla \phi(\bm{y}).
\end{displaymath}
Note that the last term represents the derivative of $\phi$ in the direction of $x - y$. Examples of Bregman divergences include the squared Euclidean distance, parameterized by $\phi(\bm{x}) = \frac{1}{2} \|\bm{x}\|^2_2$; the KL-divergence, parameterized by $\phi(\bm{x}) = \sum_i x_i \log x_i$; and the Itakura-Saito distance, parameterized by $\phi(\bm{x}) = -\sum_i \log x_i$.

Bregman divergences arise in many settings in machine learning and related areas.  In the study of exponential family distributions, there is a bijection between the class of regular Bregman divergences and regular exponential families (see~\citet{banerjee}).  In optimization, Bregman divergences arise frequently; for instance, mirror descent utilizes Bregman divergences, and Bregman divergences were originally proposed as part of constrained optimization~\cite{bregman_67}.  In the study of clustering, Bregman divergences offer a straightforward way to extend the k-means algorithm beyond the use of the squared Euclidean distance~\cite{banerjee}.  A consequence of this is a way to cluster multivariate Gaussians in a k-means framework~\cite{davis_clustering}; we will use this algorithm as a baseline later in the paper.

More recently,~\citet{functional_bd} proposed and studied an extension to standard Bregman divergences called \textit{functional Bregman divergences}, where instead of vector inputs, we compute a divergence between pairs of functions (or distributions).  In this case, given two functions $p$ and $q$, and a strictly convex functional $\phi$ whose input space is a convex set of functions and whose output is in $\mathbb{R}$, the corresponding Bregman divergence is
\begin{displaymath}
D_{\phi}(p,q) = \phi(p) - \phi(q) - \int [p(x) - q(x)] \delta \phi(q)(x) dx.
\end{displaymath}
Here $\delta \phi(q)$ is the functional derivative of $\phi$ at $q$ and the integral term calculates this derivative in the direction of $p - q$.\footnote{Note that~\citet{functional_bd} utilize the more general Fr\'{e}chet derivative.  Also, for simplicity, we limit ourselves to Riemann integrals unless otherwise noted. See Appendix for more details.}  An example of a functional Bregman divergence arises when we choose $\phi(p) = \int p(x)^2 dx$; in this case, one can work out that the functional derivative of $\phi$ at $p$ is $2p$ and that the resulting functional divergence is $\int[p(x) - q(x)]^2 dx$.

\vspace{-.2cm}
\subsection{The Symmetric Setting}
\vspace{-.2cm}
Our first goal is to relate functional Bregman divergences back to concepts in metric learning and other related learning models.  To do this, let us define a \textit{symmetric} functional Bregman divergence as a functional Bregman divergence such that $D_{\phi}(p, q) = D_{\phi}(q, p)$ for all $p$ and $q$.

Our first result characterizes the form of an arbitrary symmetric functional Bregman divergence.  This result can be stated as follows:
\begin{theorem}
A functional Bregman divergence $D_{\phi}(p,q)$ is a symmetric functional Bregman divergence if and only if it has the following form:
\begin{displaymath}
D_{\phi}(p,q) = \iint (p(x)-q(x))(p(y)-q(y)) \psi(x,y) dx dy,
\end{displaymath}
where $\psi(x,y)$ is some symmetric positive semi-definite function.
\end{theorem}

For instance, the example from above, where $\phi(p) = \int p(x)^2 dx$, can be seen as a special case where $\psi(x,y) = 1$ if $x=y$ and 0 otherwise.
The proof of the theorem appears in the appendix.  In essence, this result extends an analogous result known from the vector setting, which states that any symmetric Bregman divergence must be a Mahalanobis distance, namely $D_{\phi}(\bm{x}, \bm{y}) = (\bm{x}-\bm{y})^T A (\bm{x} - \bm{y})$ for some positive semi-definite matrix $A$~\cite{bauschke_borwein}.

Next we must show that, for particular choices of the symmetric positive semi-definite function $\psi$, as well as restrictions on $p$ and $q$, the resulting divergence yields familiar forms.  
\\

\noindent \textbf{Deep Metric Learning and Moment-Matching.}
Let us consider $\psi(x, y) = f_W(x)^T f_W(y)$, where $f_W(x)$ is an embedding given by a neural network parameterized by weights $W$.  This is clearly a positive semi-definite function, as it is an inner product between embedded data points.  Further, assume $p$ and $q$ are distributions.

First, from Fubini's theorem, observe that we can re-write the functional Bregman divergence in this special case as
\begin{displaymath}
D_{\phi}(p,q) = \| \mathbb{E}_p [f_W] - \mathbb{E}_q [f_W]\|^2.
\end{displaymath}

This is a moment-matching type of metric.  Note the similarity to the Wasserstein distance~\cite{arjovsky_arxiv17} and the maximum mean discrepancy~\cite{Gretton2012}.  (In those cases, one further takes a supremum over the function $f_W$, which we would also do when performing optimization to learn $f_W$.)

This general form of the divergence is typically difficult to compute.  We can consider the case when we have finite samples or, equivalently, we let $p$ and $q$ be given by empirical distributions over sets of points $P$ and $Q$, respectively.  In this case, the resulting divergence simplifies to
\begin{displaymath}
D_{\phi}(p, q) = \bigg \| \frac{1}{|P|} \sum_{x \in P} f_W(x) - \frac{1}{|Q|} \sum_{y \in Q} f_W(y) \bigg \|^2.
\end{displaymath}
This yields a divergence measure between distributions $p$ and $q$ that matches the first moment, similar to how MMD operates.

To make connections to deep metric learning, consider the case where $P$ and $Q$ are of size one, namely Dirac delta functions at points $x$ and $y$, respectively.  Then the divergence is simply
\begin{displaymath}
D_{\phi}(p, q) = \|f_W(x) - f_W(y)\|^2,
\end{displaymath}
or just the squared Euclidean distance after embedding the data via a neural network.  This form is precisely what nearly all deep metric learning methods employ: they learn a neural network to embed data, apply the (squared) Euclidean distance in the mapped space, and then apply a loss function such as a contrastive or triplet loss on top of this mapped distance~\cite{chopra_cvpr05,hoffer2015deep}.

\noindent \textbf{Linear Metric Learning.}
If we replace the integral in the functional Bregman divergence with a Lebesgue integral (as it was defined in the original functional Bregman divergence paper), then use the counting measure for integration, the integral in the functional Bregman divergence simply becomes a sum over the elements in the measure space.  In this case, $\psi(x,y)$ is then replaced by a positive semi-definite matrix $A$, and function inputs to the divergence are replaced by vectors $\bm{x}$ and $\bm{y}$.  Then the resulting divergence is the usual Mahalanobis distance
\begin{displaymath}
D_{\phi}(\bm{x},\bm{y}) = (\bm{x} - \bm{y})^T A (\bm{x} - \bm{y}).
\end{displaymath}
Thus, we can recover the usual Mahalanobis metric used in linear metric learning under our framework.
\\

\noindent \textbf{Kernel Metric Learning.}
We can also recover familiar kernel forms of the preceding functions.  In the case of a kernel function $\psi(x, y) = \kappa(x,y)$, the divergence recovers the moment-matching objective but with the norm induced by the kernel's reproducing kernel Hilbert space, similar to kernel MMD~\cite{Gretton2012}.  Further, in the case of a kernel function $\kappa(x, y) = g(x)^T A g(x)$, where $g(x)$ is an embedding to a reproducing kernel Hilbert space, and $A$ is a positive-definite operator, the resulting divergence in the single-sample case yields the divergence studied for Mahalanobis metric learning in kernel space~\cite{kulis_jmlr09}.

\begin{table*}[t] \label{tab:symmetric_summary}
\centering
\begin{tabular}{lllll}
\hline
Case & Integral Setting & $\psi(x,y)$ & Inputs to $D_{\phi}$ & $D_{\phi}$\\
\hline \hline
Mahalanobis Distance & Lebesgue + Count. Meas. &   $ A \succeq 0$ & Vectors $\bm{x}, \bm{y}$ & $(\bm{x} - \bm{y})^T A (\bm{x}-\bm{y})$\\
\hline
Deep Metric Learning & Riemann & $f_W(\bm{x})^T f_W(\bm{y})$ & Dirac Deltas at $\bm{x}, \bm{y}$ & $\|f_W(\bm{x}) - f_W(\bm{y})\|^2$\\
\hline
Moment Matching & Riemann & $f_W(\bm{x})^T f_W(\bm{y})$ & Distributions $p, q$ &  $\| \mathbb{E}_p [f_W] - \mathbb{E}_q [f_W]\|^2$\\
\hline
\end{tabular}
\caption{Some of the special cases of $D_{\phi}(p,q)$ for the symmetric divergence setting.}
\end{table*}

A summary of some of the special cases described in this section appear in Table~\ref{tab:symmetric_summary}.

\vspace{-.2cm}
\subsection{The General Setting}
\vspace{-.2cm}
Next we consider the more general setting, i.e., when the functional divergence may not be symmetric.  Here our goal is to introduce a parameterization of the functional divergences that are amenable to learning via neural networks.  We term the resulting divergences as \textit{deep Bregman divergences}.

A key insight of~\citet{siahkamari_arxiv} was that one can approximate a strictly convex function arbitrarily well with a piecewise linear function.  In particular, they chose to parameterize the generating function $\phi$ of a vector Bregman divergence by the following max-affine function:
\begin{displaymath}
\phi(\bm{x}) = \max_c (\bm{x}^T \bm{w}_c + b_c).
\end{displaymath}
Here $c$ ranges from $1$ to $K$, where $K$ is the number of hyperplanes used to approximate the underlying strictly convex function.  Such functions can be used to approximate any vector Bregman divergence arbitrarily well.
Thus, learning a Bregman divergence amounts to learning the weights $\bm{w}_i$ and biases $b_i$ given appropriate supervision. 

We can perform an analogous parameterization in the functional divergence setting.  By generalizing the piecewise linear functions of Siahkamari et al, we can define a convex generating functional.  The following theorem demonstrates that every convex generating functional can be expressed in terms of linear functionals, thus justifying our choice of parameterization:

\begin{theorem}
 Let $\phi(p)$ be a convex generating functional corresponding to a functional Bregman divergence $D_\phi$. Then $\phi$ can be formulated as
\begin{displaymath}
    \phi(p) = \sup_{(w,b_w) \in A} \int p(x)w(x) dx + b_w,
    \label{eqn:prop}
\end{displaymath}
\end{theorem}
 where $A$ is a set of linear functionals in which each member is characterized by $w$ and $b_w$. 
 
 For our parameterization, we replace supremum with maximum, and denote each function pair as $(w_c, b_c)$ in a countable set of functionals $A$.
 See Appendix A.2 and B for the proof and details. In the case where $p$ and $q$ are distributions, we may write this more succinctly as $\phi(p) = \max \big(\mathbb{E}_p[w_c] + b_c\big)$, where the expectation is taken with respect to the subscript distribution $p$.  Note that straightforward application of the calculus of variations reveals that the functional derivative of $\phi(q)$ is simply $w_{q^*}$, where $q^* = \mbox{argmax}_c [\int q(x) w_c(x) dx + b_c ].$
Consequently, the functional Bregman divergence between $p$ and $q$ can be expressed as $D_{\phi}(p,q) = $

\vspace{-0.1in}
{\footnotesize
\begin{equation}
\bigg (\int p(x) w_{p^*}(x) dx + b_{p^*} \bigg ) - \bigg ( \int p(x) w_{q^*}(x) dx + b_{q^*} \bigg ).
\label{functionalBD}
\end{equation}
}
\vspace{-0.1in}

For distributions, this is more succinctly $D_{\phi}(p,q) = (\mathbb{E}_p[w_{p^*}] + b_{p^*}) - (\mathbb{E}_p[w_{q^*}] + b_{q^*}).$

This parameterization of the functional $\phi$ is now amenable to learning a functional divergence given data.  In particular, we now parameterize a divergence by the corresponding weight functions $w_1, ..., w_K$ and biases $b_1, ..., b_K$.  If we assume that each of these weight functions are given by deep neural networks, then it becomes natural to set up learning problems where we aim to learn the underlying divergence given data.  The resulting deep Bregman divergences will be shown to yield novel learning problems and strong empirical performance on benchmark metric learning tasks.  In the next section we will detail our approach to extend deep metric learning to this setting.

\vspace{-.2cm}
\section{Learning Problems and Applications}
\vspace{-.2cm}
In the previous section, we saw in the symmetric setting how different choices of the functions related to a functional Bregman divergence yield existing forms, as well as how one may parameterize a general asymmetric functional Bregman divergence using deep neural networks.  Now we connect the divergences discussed in the previous section to particular learning problems.  In particular, we describe several novel applications and learning problems that arise from learning deep Bregman divergences. 

\vspace{-.2cm}
\subsection{From Deep Metric Learning to Deep Divergence Learning}
\vspace{-.2cm}

\begin{figure}[t] 
\centering
\includegraphics[width=8.2cm]{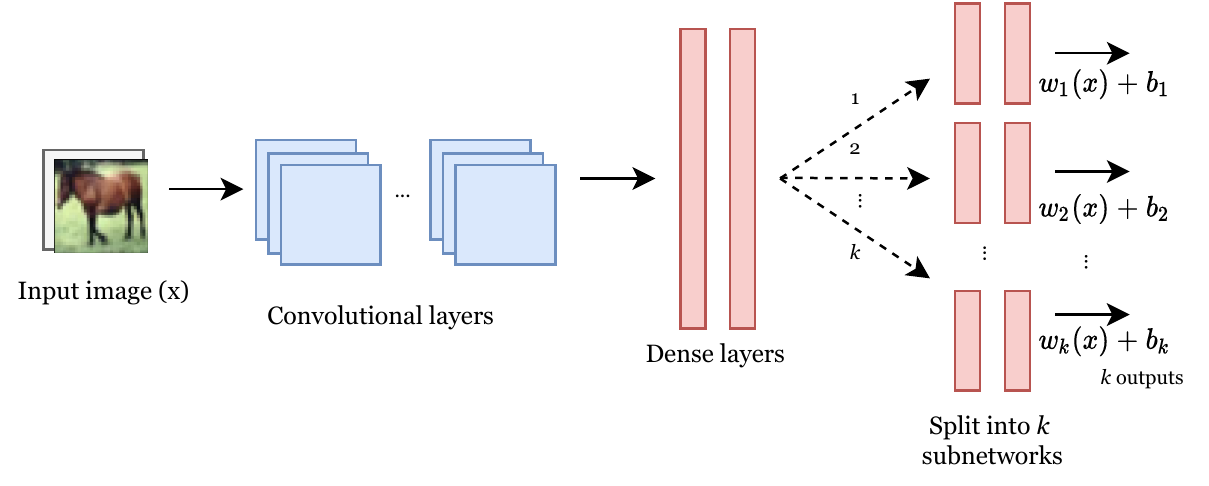}
\caption{The general architecture we employ for deep Bregman divergences on image data.  For $K$ functionals, we produce $K$ separate outputs, which are then used to compute the divergence over pairs of inputs.}
\label{fig:architecture}
\end{figure}

Consider a learning problem where we aim to learn a deep divergence given supervised data.  As with deep metric learning, we will consider the case when $p$ and $q$ are empirical distributions over single points $x$ and $y$, respectively.  We saw in the previous section that we will parameterize our deep divergence by weight functions $w_1, ..., w_K$ and biases $b_1, ..., b_K$.  To make things simpler, let us encompass all of these weight functions into a single large neural network with weights $W$.  The network will have $K$ different outputs, one for each weight function.  Many possible architectures are possible to capture this type of network; we consider an architecture where several layers are shared in the network, and then the network branches into $K$ subnetworks, each with its own independent set of weights.  See Figure~\ref{fig:architecture} for the network that we employ in our benchmark experiments.

Now, suppose we pass $x$ through the network.  Each subnetwork $c$ produces a single output $w_c(x) + b_c$, and there are $K$ total outputs, one per subnetwork.  The index of the maximum output is $p^*$.  Similarly, pass $y$ through the network; the index of the maximum output across the $K$ subnetworks is $q^*$.  Then, by~\eqref{functionalBD}, the divergence is the difference between the output of $x$ at $p^*$ and the output of $x$ at $q^*$.
For instance, suppose that each of the $K$ outputs corresponds to a different class.  Then the divergence will be zero if both points achieve a maximum value for the same class (i.e., they are both classified into the same class).  The divergence is non-zero if the points are assigned to different classes, and the divergence grows as the two outputs become more disparate.

One can now set up a divergence learning problem over pairs or triples of points under this framework.  Suppose we are given triples of points ($x$, $y$, $z$), where $x$ should have a smaller divergence to $y$ than to $z$.  One can easily apply existing loss deep metric learning loss functions---the triplet loss or contrastive loss are the two most common ones---with the learned divergence in place of the usual squared Euclidean distance.  See the appendix for definitions of standard loss functions for deep metric learning.  In experiments, we will compare existing deep metric learning approaches to the more general deep divergence learning problem considered here, and we will see that we obtain gains over the existing models on standard benchmarks.

\vspace{-.2cm}
\subsection{Learning over Distributions}
\vspace{-.2cm}
A key advantage to our framework is that we need not restrict ourselves only to divergences between single points.  As we saw earlier, we can also capture divergences between distributions of points that are similar to what is used for the MMD and the Wasserstein distance.  Here we will discuss applications involving learning divergences over distributions.

\noindent \textbf{Data Generation.} Consider the problem encoutered in many GAN applications: we aim to learn a generator for data such that we minimize some distributional divergence between the real and generated data distributions.  In existing GAN literature, divergences considered include the Jensen-Shannon divergence~\cite{gans}, MMD distance~\cite{gmnn,mmd_gan}, and the Wasserstein distance~\cite{arjovsky_arxiv17}.

Under the deep divergence framework, rather than employing a fixed divergence, we can learn one from data.  In this setting, we consider two distributions $p_{synth}$ and $p_{real}$, corresponding to distributions of generated and real data, respectively.  Assume that $p_{synth}$ is generated by passing randomly-generated input data through a generator $g$, as is standard with GAN models.  As with GAN training, learning proceeds in an adversarial manner.  We aim to learn a generator to minimize $D_{\phi}(p_{synth}, p_{real})$, while simultaneously we aim to learn weights of the underlying network parameterizing $D_{\phi}$ to maximize $D_{\phi}(p_{synth}, p_{real})$.  As with GANs, we alternate between gradient updates for these two objectives.

We note that, in practice, it is useful to restrict our attention to the case when $K=2$, as it yields a particularly interpretable model.  In this case, we can think of one of the two subnetworks as outputting a larger value on real data, while the other subnetwork as outputting a larger value on synthetic data.  Thus, the network that parameterizes the divergence is analogous to the discriminator in a GAN model.  When training the underlying weights of this network $W$, we can take pairs or triples of real and synthetic data and utilize a triplet or contrastive loss to encourage the output on the real data to be larger for one subnetwork and the output on the synthetic data to be larger for the other subnetwork.  Similarly, when training the generator $g$, we use a loss that encourages real and synthetic data to both have the same maximal output.

\noindent \textbf{Semi-Supervised Distributional Clustering.}
As another application of learning divergences over distributions, consider a scenario where instead of clustering a set of data points, we aim to cluster a set of distributions.  In this setup, each distribution may correspond to an empirical distribution over a set of points---for instance, we may have a distribution of ratings for each item in an online store.  The goal is: given a set of such distributions, to cluster the distributions together into a set of clusters.

\citet{davis_clustering} considered a version of this problem where each distribution was given by a multivariate Gaussian.  Since the KL-divergence between multivariate Gaussians is itself a Bregman divergence, one can use properties of Bregman divergences to generalize the k-means algorithm to this setting.  Here, we will consider a version of this problem that is both semi-supervised (so pairs of distributions that should or should not be clustered together are provided over a training set), and does not assume that each distribution is a multivariate Gaussian. Our approach also removes the implicit assumption that the means of the distributions are linearly separable for each cluster. 

Analogous to Davis and Dhillon, given a functional Bregman divergence defined over distributions, one can apply a generalization of k-means to cluster the distributions.  As shown by~\citet{functional_bd}, the mean minimizes the expected functional Bregman divergence over a set of distributions, analogous to the finite-dimensional case.  Thus, k-means can be generalized to a setting where the squared Euclidean distance between vectors is replaced by the corresponding functional Bregman divergence over distributions.

If we represent each distribution by an empirical distribution over its underlying points, we can easily compute a parameterized functional Bregman divergence between pairs of distributions.  In our experiments, we will consider in particular learning a symmetric divergence on supervised data using the moment-matching distance with a contrastive or triplet loss.  Then, once we have learned the divergence from data, we replace the squared Euclidean distance in the k-means algorithm with the learned divergence to directly cluster data in the test set.

\begin{figure*}[t!]
\minipage{0.2\textwidth}
    \includegraphics[width=\linewidth]{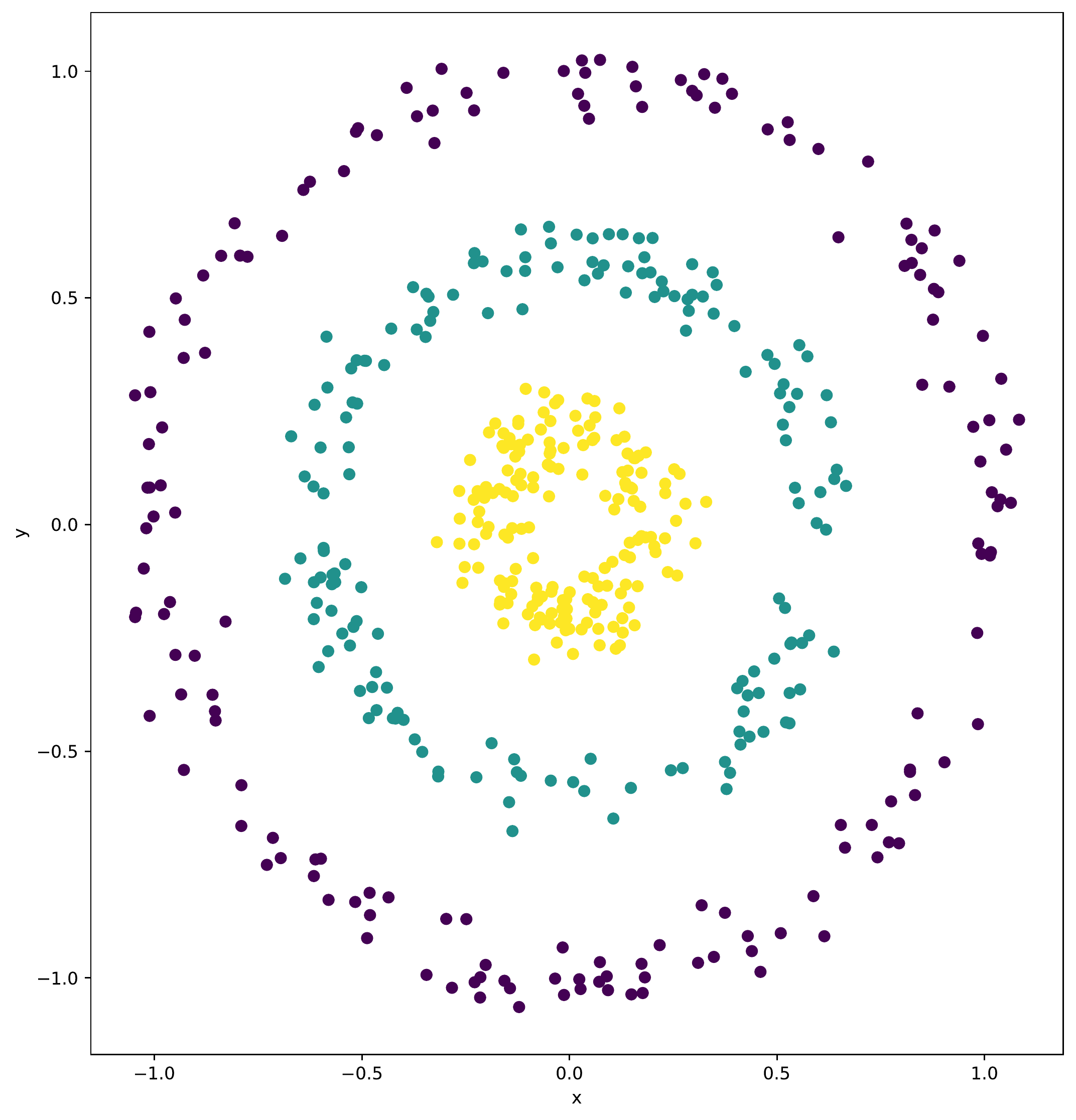}
\endminipage \hfill
\minipage{0.2\textwidth}

  \includegraphics[width=\linewidth]{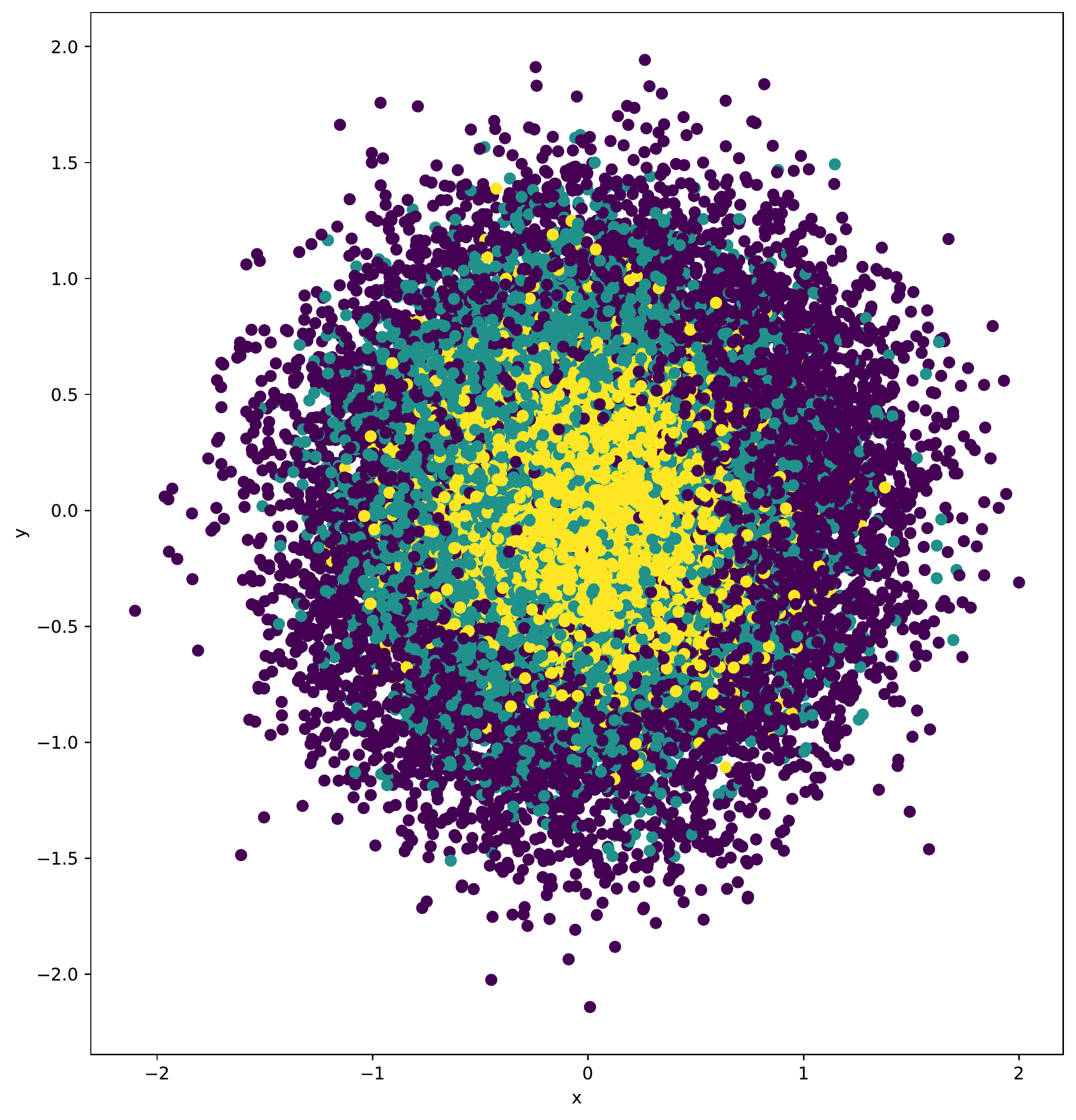}
\endminipage\hfill
\minipage{0.2\textwidth}
  \includegraphics[width=\linewidth]{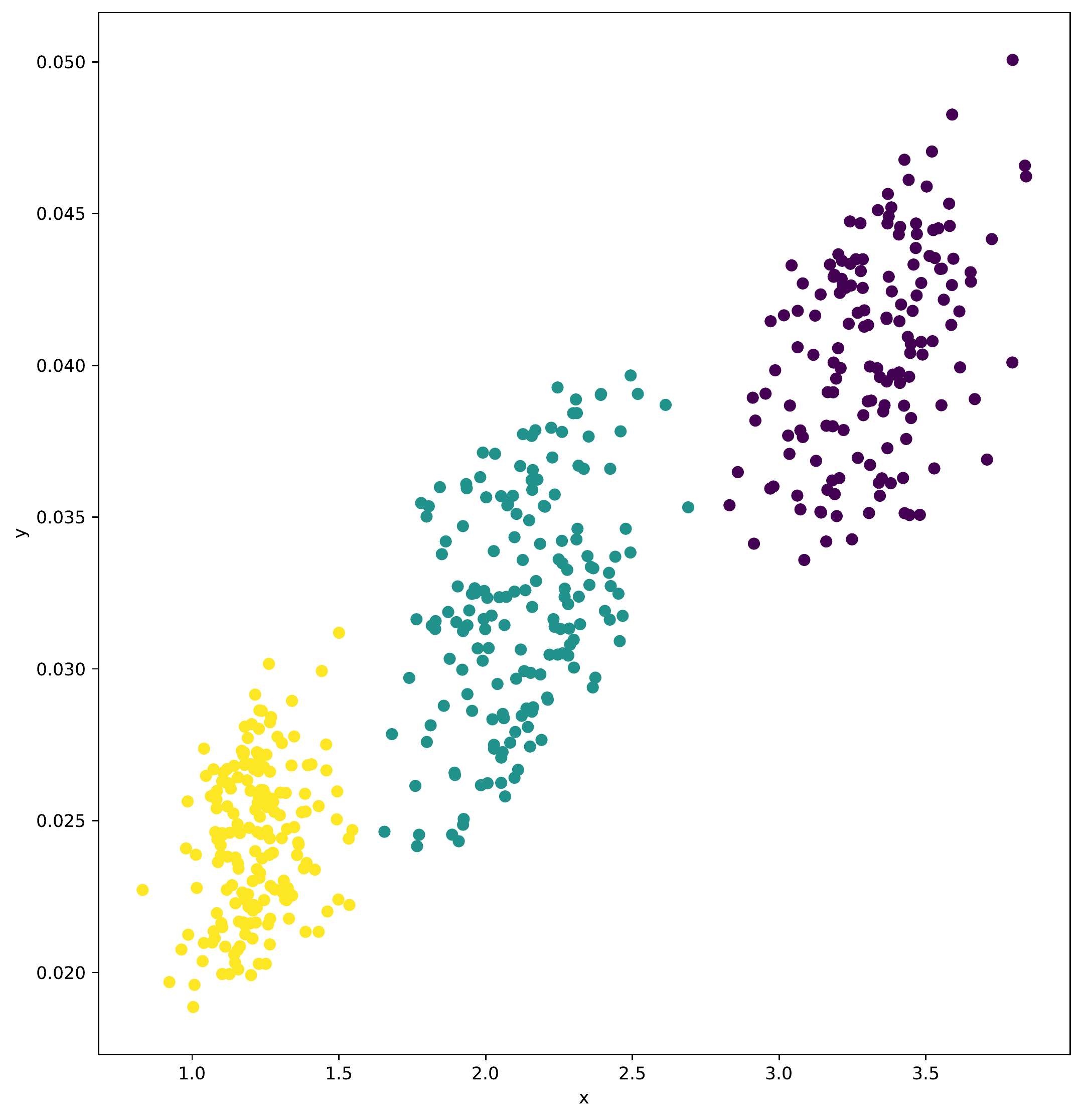}
\endminipage\hfill
\minipage{0.2\textwidth}%
  \includegraphics[width=\linewidth]{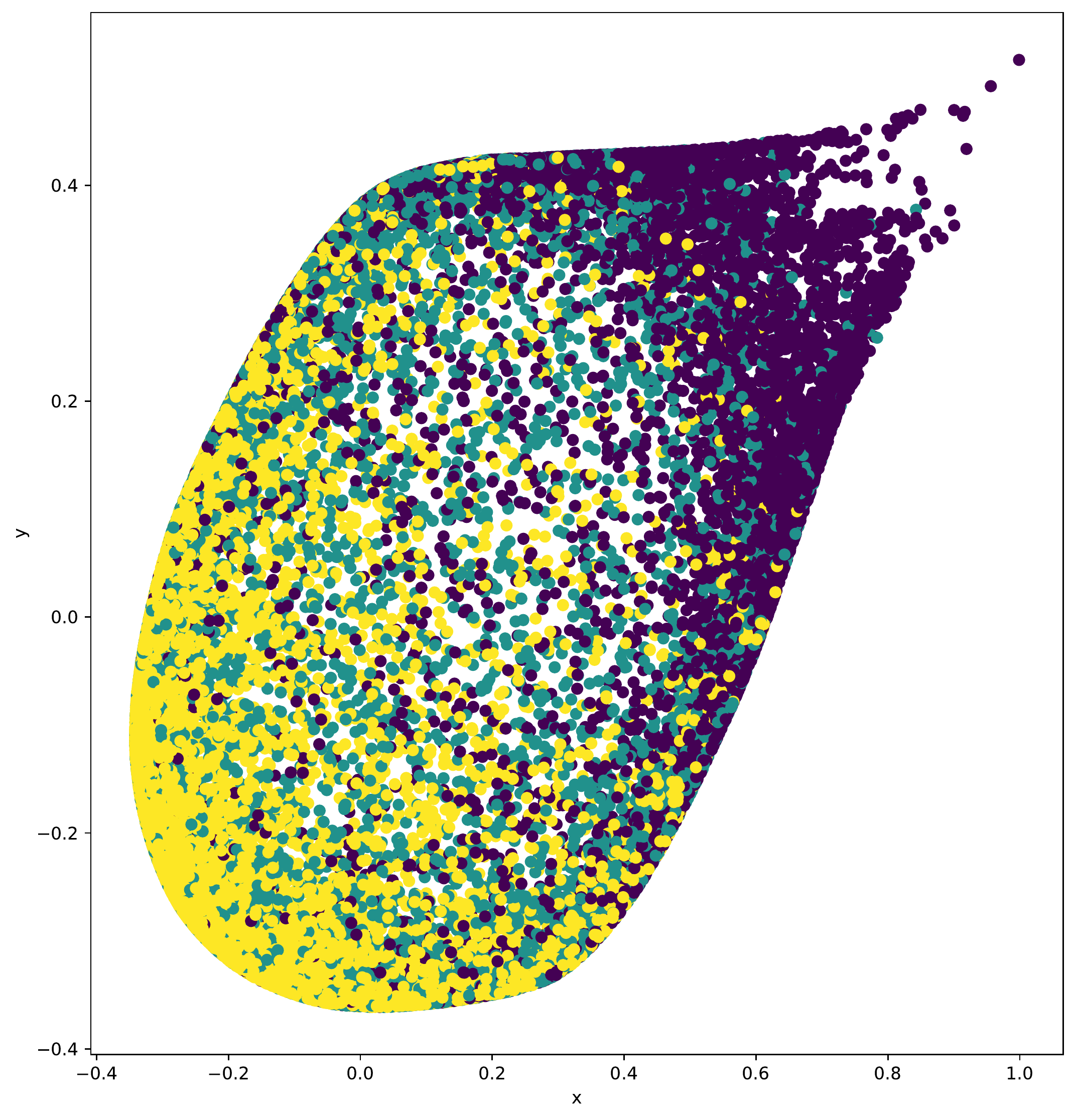}
\endminipage
\caption{{  (Left) Plot of the means of the $n=500$ Gaussian distributions, color-coded by cluster identity.  (Middle left) Plot of data after generating 50 points from each Gaussian.  (Middle right) Embedding learned by our method using contrastive loss with a moment-matching function.  (Right) Embedding learned by the baseline deep learning approach.}}
\label{fig:dataplots}
\end{figure*}

\vspace{-.2cm}
\section{Experimental Results}
\vspace{-.2cm}

We now empirically compare our proposed deep divergence framework to existing models.  Due to space considerations, some further details and results are available in the supplementary material.

\vspace{-.2cm}
\subsection{Clustering}
\vspace{-.2cm}
To begin, we consider a simple demonstration of the advantages of our approach on synthetic data for the semi-supervised distributional clustering problem.  We generated $n = 500$ training points, each assigned to one of three clusters.  Each data point is represented by a multivariate Gaussian; the means of these Gaussians were uniformly sampled over rings of radius $.2, .6$, and $1$ plus Gaussian noise, depending on the cluster identity, and the covariance of each Gaussian was $.1$ times the identity. See Figure~\ref{fig:dataplots} for a plot of sampled means, along with data after generating from these Gaussians.  We also generated $n=200$ test points in the same manner.



\begin{table}[t]
\centering
\resizebox{1\columnwidth}{!}{%
\renewcommand{\arraystretch}{1.4}
\begin{tabular}{clcccccc}
\hline
\multicolumn{3}{c}{\multirow{2}{*}{\textbf{Metrics}}} & \multicolumn{2}{c}{Baseline Method} & \multicolumn{2}{c}{Our Method} & \multirow{2}{*}{\begin{tabular}[c]{@{}c@{}}David \& \\  Dhillon\end{tabular}} \\
\multicolumn{3}{c}{} & Triplet & Contrastive & Triplet & Contrastive &  \\ \hline \hline
\multicolumn{2}{c}{\multirow{2}{*}{RI}} & Mean & 0.638 & 0.639 & \textbf{0.997} & \textbf{0.999} & 0.550 \\
\multicolumn{2}{c}{} & Std & 0.005 & 0.005 & 0.003 & 0.003 & 0.009 \\
\multicolumn{2}{c}{\multirow{2}{*}{ARI}} & Mean & 0.197 & 0.198 & \textbf{0.993} & \textbf{0.997} & 0.005 \\
\multicolumn{2}{c}{} & Std & 0.012 & 0.013 & 0.007 & 0.006 & 0.012 \\ \hline
\end{tabular}
}
\caption{Rand index and adjusted rand index scores for different clustering experiments, where the baseline method treats each training point independently.} 
\label{tab:clustering}
\end{table}

We compare three approaches to cluster the data.  Our first baseline is the method of~\citet{davis_clustering}, which is an unsupervised clustering algorithm designed specifically to cluster multivariate Gaussian distributions.  The second baseline applies deep metric learning on all generated points from all the Gaussians; we apply contrastive and triplet losses separately and learn a 3-layer multilayer perceptron (MLP) over the data in each case.  The number of units in each layer were set to 1000, 500, and 2, and standard ReLU activation was used.  The third approach is our method; we apply the (empirical) moment-matching function from the symmetric setting, treating each distribution as its own data point, in conjunction with a contrastive and triplet losses to learn a 3-layer MLP with the same settings as the baseline MLP.  On the test set, we use the learned divergence in place of the squared Euclidean distance in a k-means algorithm for both the second and third method.

We compute the rand index and adjusted rand index scores on the test set in each case, averaged over 10 runs for each of the three methods.  The results are given in Table~\ref{tab:clustering}.  The Davis \& Dhillon method cannot cluster the multivariate Gaussians, as their method is restricted to linear separability of the means.  The baseline deep metric learning method fails due to the overlap of the generated data across clusters, whereas the distributional divergence approach is able to perfectly cluster the test data in most runs.  We can also visualize the embeddings learned by the second and third method, where we see that our learned embeddings capture the correct cluster structure, as pictured in Figure~\ref{fig:dataplots}.

Further experiments were performed on real datasets, the results of which are enumerated in Appendix C.

\begin{table}[b]
\centering
\resizebox{.95\columnwidth}{!}{%
\renewcommand{\arraystretch}{1.3}
\begin{tabular}{ccccc}
\hlineB{1.75} \multirow{2}{*}{\textbf{Datasets}} &
  \multicolumn{2}{c}{Euclidean} & \multicolumn{2}{c}{Deep Bregman} \\ 
& Triplet  & Contrastive &  Triplet & Contrastive \\ 
\hline \hline
MNIST & 99.50 & \textbf{99.63} & \textbf{99.61} & 99.56 \\ 
Fashion MNIST & 93.24  & 93.57  & \textbf{94.90} & \textbf{94.00} \\ 
SVHN & 92.58 & \textbf{94.88} & \textbf{94.03} & 94.12 \\
Cifar10 & 77.00 & 79.40 & \textbf{81.40} & \textbf{80.80} \\ 
STL10 & 59.97 & \textbf{63.10} & \textbf{62.64} & 60.91\\ \hlineB{1.75}
\end{tabular} 
}
\caption{K-nn classification accuracy results on the given datasets (without data augmentation or using learned features).  The bold values indicate the best triplet loss (Bregman versus Euclidean) and contrastive loss (Bregman versus Euclidean) results. } 
\label{tab:benchmark}
\end{table}  

\vspace{-.2cm}
\subsection{Deep Metric Learning Comparisons}
\vspace{-.2cm}

\begin{table}[t]
\resizebox{.95\columnwidth}{!}{
\renewcommand{\arraystretch}{1.3}
\begin{tabular}{lclc}
\hlineB{1.75}
\multicolumn{2}{c}{\textbf{Model hyperparams}} & \multicolumn{2}{c}{\textbf{Training hyperparams}} \\ \hline \hline
layers                & 2 - 5                & margin                     & 0.1 - 2.0               \\
conv filters                & 16 - 128              & epochs               & 10 - 40                 \\
conv kernels                & 3 - 9                & learning rate              & $10^{-5}$ - $10^{-1}$             \\
conv biases                   & T / F                & batch size                 & 32-128                \\
poolings                     & T / F                & optimizer                  & adam / sgd / rms          \\
batchnorms                   & T / F                & K in k-nn                        & 5 - 10                  \\
dense units              & 50 - 300             & normalization        & T / F                \\ \hlineB{1.75}
\end{tabular}
}
\caption{Hyperparameter intervals used for tuning. First 100 iterations are used to narrow down the space, then 200 more iterations are run for each benchmark. T: True, F: False. }
\label{tab:hyper}
\end{table}

Next we consider comparisons between our general deep divergence learning framework and existing deep metric learning models on standard benchmarks, to demonstrate that our approach's flexibility yields improved performance on several datasets and tasks.

We compare standard deep metric learning approaches to our proposed approach on the four benchmark datasets used in the original triplet loss paper~\cite{hoffer2015deep}---MNIST, Cifar10, SVHN, and STL10---as well as Fashion MNIST.  We use the same basic architecture for the deep Bregman divergence network as shown in Figure~\ref{fig:architecture}; for the Euclidean case we do not employ separate subnetworks in the dense layers.  We treat several architecture choices as hyperparameters and validate over these hyperparameters using Bayesian optimization (tuned separately for each dataset); Table~\ref{tab:hyper} lists the hyperparameters that we search over, along with the ranges of values considered.

We consider separately both triplet loss and contrastive loss, and report in bold the best values for each loss.  For the triplet loss, we consider all triplets in a batch when computing the loss.  We perform no data augmentation.
Results are shown in Table~\ref{tab:benchmark}, where we see small but significant gains in classification accuracy for the Bregman method as compared to the standard deep metric learning approach, particularly in the triplet loss case.  On Fashion MNIST, we outperform the current state-of-the-art for no data augmentation (94.23\% from~\citet{denser}), even though we are not directly training a classifier.  We also note that we would expect further gains in performance with more sophisticated architectures (e.g., ResNets and other more recent architectures), perhaps yielding near state-of-the-art performance on more datasets; however, the main goal of this comparison is not to achieve state-of-the-art performance but rather to present a fair comparison between the Bregman and Euclidean approaches on standard benchmarks.

\begin{figure}[t]
\minipage{0.23\textwidth}
    \includegraphics[width=\linewidth]{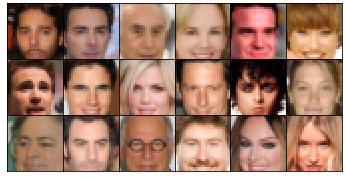}
\endminipage \hfill
\minipage{0.23\textwidth}
  \includegraphics[width=\linewidth]{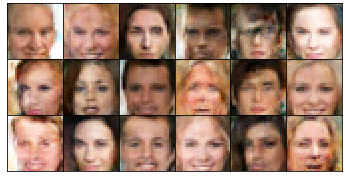}
\endminipage\hfill
\vskip\baselineskip
\minipage{0.23\textwidth}
  \includegraphics[width=\linewidth]{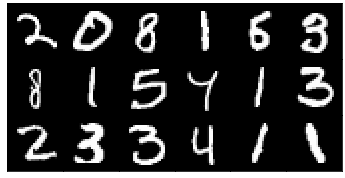}
\endminipage\hfill
\minipage{0.23\textwidth}%
  \includegraphics[width=\linewidth]{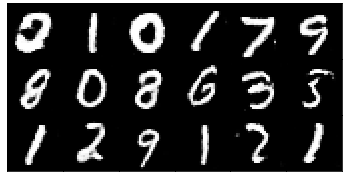}
\endminipage
\caption{ Real and generated sample batches from CelebA (Top row) and MNIST (Bottom row) datasets. } 
\label{fig:gan}
\end{figure}

\vspace{-.2cm}
\subsection{Unsupervised Data Generation}
\vspace{-.2cm}
Finally, we consider some qualitative results where we show that our approach can be used for generating data with similar performance to GANs.  We consider the problem discussed earlier, namely where we train a deep divergence model to minimize a learned divergence between real and synthetic data.  We apply our approach on 28x28 MNIST and CELEBA datasets, as is standard for GAN applications. We adjust the strides to adapt the network for different input sizes. We keep model structures close to standard in order to show the effectiveness of the divergence formula we introduced. We use a generator consisting of 4 deconvolutional layers and a discriminator (i.e., the network parameterizing the deep Bregman divergence) with 4 convolutional layers, with a dropout rate of 0.5 in between the layers as well as lrelu and tanh activations. In the discriminator network, the convolutional layers are followed by two 2-layer subnetworks (again similar to Figure \ref{fig:architecture}, where $K=2$ in this case). 
For the discriminator, we use the contrastive loss with a margin of 0.4, whereas the generator directly attempts to minimize deep Bregman divergence between the real and generated images. More hyperparameter details are given in the appendix. 

Some randomly chosen results are presented in Figure \ref{fig:gan}, where we see that the distribution divergence learned by our method is able to generate realistic-looking images with no labeled supervision. We note that further theoretical analysis and experimentation of these methods is required to determine situations where our loss functions may be more desirable than existing GAN approaches.

\vspace{-.2cm}
\section{Conclusions}
\vspace{-.2cm}
In this paper, we examined a novel generalization of both Bregman divergence learning and deep metric learning, which we call deep divergence learning.  This framework offers several appealing advantages: it unifies a number of existing ideas in metric learning under a single framework, it suggests a way to extend deep metric learning beyond the Euclidean setting, and it naturally yields learning problems involving divergences over distributions.  Empirically we have seen advantages of our approach compared to existing deep metric learning methods.

\bibliographystyle{icml2020}
\bibliography{hugebib}

\begin{appendices}
\section{Notation and Definitions}

In this section, we first briefly define triplet and contrastive losses used in the main paper. Then, we introduce basic concepts from functional analysis and the notation used for extending vector spaces to function spaces, which will be used for our proofs.

\subsection{Definitions of contrastive and triplet losses}

As a reminder for the reader, we provide the definitions of contrastive and triplet losses. The idea behind these losses is to enforce a small distance between similar inputs, and a large distance for dissimilar inputs. The Euclidean distance is utilized as the distance measure, denoted by $d$.

\textbf{Contrastive loss}. The contrastive loss takes an input pair, $x_a$ and $x_b$, together with a relationship label $y$ that takes the value of 1 if the inputs are similar and 0 otherwise. The loss function for a single $x_a$, $x_b$ pair is:
\begin{displaymath}
\mathcal{L}(x_a,x_b) = y d(x_a,x_b) + (1 - y) \max \{m - d(x_a, x_b) , 0 \}^2,
\label{eqn:contrastive}
\end{displaymath}
where $m$ is a margin value to separate dissimilar samples, chosen as a hyperparameter.

\textbf{Triplet loss}. The triplet loss takes an input triplet $x_a. x_p, x_n$, where the anchor $x_a$ has one similar input $x_p$ and one dissimilar input $x_n$. The loss function for a single triplet is:
\begin{displaymath}
\mathcal{L}(x_a,x_p,x_n)  = \max \{d(x_p, x_a) - d(x_n, x_a) + m, 0\},
\label{eqn:triplet}
\end{displaymath}
where $m$ is again a margin value to separate relative distances of the similar and dissimilar pairs.

Typically, the distance measure $d$ in both loss functions is the Euclidean distance; however, for our loss functions, we replace the distance measure by our learned deep Bregman divergence.

\subsection{Assumptions and definitions from functional analysis}

 We first present basic notation from functional analysis, since we extend vector spaces to function spaces to derive this formulation. 

 Assume we have a finite  measure space $(\chi, \Sigma, \mu)$ which is Lebesgue-measurable, and $\chi \in \mathbb{R}^d$.  Note that we mainly consider a set of distributions in this paper, which is a special case that uses a Radon measure and a bounded Borel set, but we continue with the more general case. Consider a set of measurable functions $F \subseteq L^p$, defined as $F = \{f \in F \ | \ f:\chi \rightarrow R$, $||f ||_p \leq C_1 < \infty$ \ and \ $f \geq 0 \}$, where $C_1$ is a constant and $1 \leq p \leq \infty$. The restriction that $f \geq 0$ is not limiting, since it can be easily satisfied by using its equivalence class obtained by only applying an affine transformation \cite{functional_bd}.

Assume $W \subseteq$ $L^p$ is a compact set of functions. All linear functionals have a continuous integral representation with respect to our focus of measure space \cite{gierz1987integral}, with a corresponding function $w\in W, \ w : \chi \rightarrow \mathbb{R}$. Similarly, we can characterize affine functionals by their function and constant pairs $ \textbf{A}  = \{(w, b_w) \ | \ w \in W, b_w \in \mathbb{R} \ \text{and} \ |b_w | \leq C_2 \}$, with $C_2$ a constant. 

For a convex functional $\phi$, we denote its Fr\'{e}chet derivative as $\delta \phi(p)$ and the epigraph of $\phi$ as $epi$ $\phi$; with their definitions briefly given below \cite{gelfand2000calculus} : 

\textbf{Fr\'{e}chet  derivative of $\phi$}. If for every  $h \in W$, there exists $\delta \phi(f)$ s.t.
\begin{equation*}
\lim_{||h||_p \rightarrow 0 } \frac{\phi(f + h) - \phi(f) - \delta \phi (f)[h]}{||h||_p}  = 0,
\end{equation*}
then $\phi(f)$ is Fr\'{e}chet  differentiable and $\delta \phi(f)$ is the Fr\'{e}chet  derivative of $\phi$ at $f$. 

\textbf{Directional Fr\'{e}chet  derivative of $\phi$}. The derivative of a functional $\phi$ at $f$ in the direction of a function $g$ is defined as:

\begin{equation*}
\delta \phi [f; g] = \int \delta \phi (f)(x) g(x) dx.
\end{equation*}

\textbf{Epigraph of $\phi$}. The epigraph of a functional $\phi$ is defined as:
\begin{equation*}
epi \  \phi  := \{ (f, c ) \in F\times \mathbb{R} \ | \ \phi(f) \leq c\}.      
\end{equation*}

\section{Proof of Theorem 3.1}
\begin{proof} To prove the result, we can generalize a known symmetry result for standard Bregman divergences seen in Bauschke \& Borwein, Lemma 3.16~\cite{bauschke_borwein}, or this Mathematics Stack Exchange discussion\footnote{https://math.stackexchange.com/questions/2242980/bregman-divergence-symmetric-iff-function-is-quadratic}.

We start by establishing that any symmetric functional Bregman divergence has the form given in the statement of the theorem.  Let $0_f$ be the zero-function (given, for example by the function $p-p$ for any $p$).  We can assume without loss of generality that $\phi(0_f) = 0$ and $\delta \phi(0_f) = 0$---we can always add a constant to $\phi$ to ensure the first property, and we can subtract $\int p(x) \delta \phi(0_f) dx$ from $\phi$ to ensure the second property, both without changing the resulting Bregman divergence.

Next, if $D_{\phi}(p,q) = D_{\phi}(q,p)$ for all $p,q$, then writing out the Bregman divergences and equating them yields
\begin{align}
& \phi(p) - \phi(q) - \int(p(x) - q(x)) \delta \phi(q)(x) dx \nonumber \\
& = \phi(q) - \phi(p) - \int(q(x)-p(x)) \delta \phi(p)(x) dx.
\label{eqn:symmBD}
\end{align}
Letting $p = 0_f$ and simplifying the above equation (and using $\phi(0_f) = 0$ and $\delta \phi(0_f) = 0$), we obtain the following:
\begin{displaymath}
2 \phi(q) = \int q(x) \delta \phi(q)(x) dx.
\end{displaymath}
Note that this equation holds for any $q$.  Plugging this equation (along with the same equation where $p$ has replaced $q$) into (\ref{eqn:symmBD}), we obtain the following identity:
\begin{equation}
\int p(x) \delta \phi(q)(x) dx = \int q(x) \delta \phi(p)(x) dx.
\label{eqn:bd_identity}
\end{equation}
This can be used to establish that $\delta \phi$ is linear.  For example, to establish that $\delta \phi$ is homogeneous, we must show that $\delta \phi(\alpha p) = \alpha \delta \phi(p)$, for non-zero $\alpha$.  Using (\ref{eqn:bd_identity}) twice (first and third line), we can establish the following for any $p$ and $q$:
\begin{eqnarray*}
\int q(z) \delta \phi(\alpha p)(z) dz & = & \int \alpha p(z) \delta \phi(q)(z) dz\\
& = & \alpha \int p(z) \delta \phi(q)(z) dz\\
& = & \alpha \int q(z) \delta \phi (p)(z) dz.
\end{eqnarray*}
This can then be used to show that $\delta \phi(\alpha p) = \alpha \delta \phi(p)$: for any point $x$, suppose $p$ is a Dirac delta function at $x$.  Then the above equation establishes that $\delta \phi(\alpha p)$ equals $\alpha \delta \phi(p)$ at $x$.  Since the equation is true for all $p$, then $\delta \phi(\alpha p)$ equals $\alpha \delta \phi(p)$ for all points. 

A similar argument can be used to establish that $\delta \phi(p + q) = \delta \phi(p) + \delta \phi(q)$.  In particular, $\int r(z) \delta \phi(p+q)(z) dz$
\begin{eqnarray*}
 & = & \int  (p(z) + q(z)) \delta \phi(r)(z)\\
& = & \int p(z) \delta \phi(r)(z) + \int q(z) \delta \phi(r)(z) dz\\
& = & \int r(z) \delta \phi(p)(z) dz + \int r(z) \delta \phi(q)(z) dz
\end{eqnarray*}
for all $r$, establishes that $  \delta \phi(p + q) = \delta \phi(p) + \delta \phi(q)$ and choosing $r$ as Dirac delta functions ensures this equality for all points.

In the case of functions, if a gradient function $\delta \phi$ is linear, then the function $\phi$ must be quadratic; this is because we take an anti-derivative of a linear function and obtain a quadratic function.  In the functional case, this means that $\phi$ must have the following form:
\begin{displaymath}
\phi(p) = \iint p(x) p(y) \psi(x,y) dx dy,
\end{displaymath}
where $\psi$ is a symmetric, positive semi-definite function.  (In the vector setting, $\phi(x) = \bm{x}^T A \bm{x}$ for a positive semi-definite matrix $A$, so this is a generalization to the functional setting.)  One can verify that the gradient $\delta \phi$ is of the form
\begin{displaymath}
\delta \phi (p)(y) = 2 \int p(x) \psi(x,y) dx,
\end{displaymath}
which is indeed a linear function.  Given this form for $\phi$, the final step is to plug $\phi$ into the definition for the functional divergence and to simplify the resulting divergence.  After simplification using the definition of $\phi$ and its derivative, along with the fact that $\psi(x,y) = \psi(y,x)$, we obtain
\begin{displaymath}
D_{\phi}(p,q) = \iint (p(x)-q(x))(p(y)-q(y)) \psi(x,y) dx dy.
\end{displaymath}

Now that we have established one direction of the theorem, we can establish the other.  This direction is considerably simpler.  We must show that a divergence that has the form
\begin{displaymath}
D_{\phi}(p,q) = \iint (p(x)-q(x))(p(y)-q(y)) \psi(x,y) dx dy
\end{displaymath}
is in fact a symmetric functional Bregman divergence.  The fact that it is symmetric follows directly.  The fact that it is a functional Bregman divergence follows from the fact that choosing the strictly convex functional $\phi(p) = \iint p(x) p(y) \psi(x,y) dx dy$ yields the resulting divergence.

\end{proof}
\section{Proof of Theorem 3.2}

In this section, we show that our convex generating functional form is justified in that any convex functional can be represented as a supremum over linear functionals.

Up to this point, we notated convex functionals as $\phi(p)$, in terms of only their input functions. Here we will use the notation $\phi(x; p(x))$ for convex functionals, where $x$ refers to the input of the function $p$. 

\begin{proof}
 \textbf{$(\supseteq)$} We first show that the right hand side is indeed a convex functional.

 We will use the standard definition of convexity since it directly extends to the functional case. The domain of the functionals is a convex subset since for all $\lambda \in [0,1]$, and $p,q \in L^p$, $ || \lambda p + (1 - \lambda q)||_p < \infty,$ so $\lambda p + (1 - \lambda q) \in L^p$ naturally.

For an arbitrary pair $(w, b_w)$, and $p,q \in F$ we have:
\begin{align*}
&\int (\lambda p(x) + (1 - \lambda) q(x)) w(x) dx + b_w \leq \\ &\lambda  \bigg [ \int p(x) w(x) dx + b_w \bigg ] +   (1 - \lambda) \bigg [ \int q(x)w(x)dx + b_w \bigg ] \\ &= \lambda \phi_p(x; p(x))+ (1 - \lambda) \phi_q(x; q(x)) \\ 
&\leq \lambda \phi^*(x;p(x)) + (1 -\lambda) \phi^*(x;q(x)), \end{align*}

where $\phi^*$ represents the supremum attained for the right-hand side of~(\ref{eqn:prop}), and $\phi_a(b) = \int b(x)a(x)dx + b_a $ . Since the inequalities hold for all $(w, b_w)$, we can take the $\sup$ of the first line and obtain:
\begin{align*}
    \phi^*(\lambda p + (1 - \lambda)q(x) ) &\leq \lambda \phi^*(x;p(x)) \\ & + (1 - \lambda) \phi^*(x;q(x)).
\end{align*}

$(\subseteq)$ We now show that for a given convex functional $\phi(x; p(x))$, we can find a set of affine functionals to write it as~(\ref{eqn:prop}).

Assume $\delta \phi(x; p (x))$ is the Frechet derivative of $\phi$ at function $p \in W$. Then $\delta \phi(x; p (x))$ is a linear operator. Since $\phi$ is continuous and bounded, we can find $r(x) = \arg \inf \delta \phi(x; p (x))$\footnote{$r(x)$ also can be constructed from the $\epsilon$-subdifferentials of $\phi$ to ensure existence. }. Define $ \delta^{'}_\phi(x; p (x)) := \delta \phi(x; p (x)) + \phi(x; r (x)) \geq 0$. This positive functional can be represented in an integral form by Riesz-Markov-Kakutani representation theorem on the measure $d(\delta^{'}_\phi(x))$ \cite{frechet1907ensembles}. Note that we can always add or substract properly scaled constant terms and preserve the information since these transformations are linear. For a given $p'$, applying Riesz theorem gives us the representation below:
\begin{align*}
    \delta \phi(x; p' (x)) = \int \delta \phi'(x; p'(x)) p'(x) dx,
\end{align*}
with a support function
\begin{align*}
    l_{\phi_{p'}}(x; p(x)) &= \int \delta \phi'(x; p'(x)) p(x) dx \\&+ \phi(x; p' (x)) -  \delta \phi(x; p' (x)).
\end{align*}

We also have ${l_{\phi_{p'}}}(x; p(x)) \leq \ epi \ \phi(x; p(x))$ for all $p, p' \in W$, since $\phi$ is a convex functional. ${l_{p'}}(x; p(x)) \leq \phi(x; p(x))$ for all $p$  since we are on a compact domain and $\phi$ is continuous and convex. Now for a given convex functional $\phi(x; p(x))$, consider $\bigcup_{p' \in W} {l_{\phi_{p'}}}(x; p(x))$ 
as a set of affine functionals, further denoted by $L_{\phi}$ for convenience. Define:
\begin{align*}
   \psi(x;p(x)) = \sup_{l \in L_\phi} \int l(x)p(x)dx + b_l.
\end{align*}
$\psi$ is a convex functional by the first part of the proof. Since $\phi$ is convex, for all $p \neq p'$ we have $\phi(x; p^{'} (x)) -  l_\phi(x; p^{'} (x)) \geq 0$, so for an arbitrary $p$, $\psi(x;p(x)) = l_\phi(x; p (x)) $. This concludes $\psi(x;p(x))$ forms a set of functionals to construct $\phi$. 
\end{proof}

Note that if we restrict our space to a set of $n$ distributions, then all we need to know is at most $n$ corresponding maximizing affine functionals; in this case the supremum can be replaced by the maximum as we did in the paper.

\section{Applications Details}

\subsection{Additional GAN Model Details}
In this section, we present more training details related to our GAN model. We use RMSprop optimizer with a momentum value of 0.99, and set the learning rates to $10^{-3}$ for the discriminator and $3 \times 10^{-3}$ for the generator. The minibatch size is chosen as 64. Our main model has convolutional layers with kernel sizes equal to 5 and filter sizes equal to 64. The strides are halved towards the final layers. Stride sizes are determined based on the input image dimensions. 

In our experiments, we incorporate contrastive loss into deep Bregman learning in order to supervise the  discriminator. However, our distributional loss formula has the potential to be directly used in the GAN setting, which we leave as a future work.

We provide the loss plots for the generator and the discriminator through the training phase below in Figure~\ref{fig:gan_plots}. We observe that the discriminator first learns the metric, then the training preserves the balance between the two networks. We note that image quality still improves for a while after the losses become saturated, due to the nature of contrastive loss.    

\begin{figure}[t]
\minipage{0.24\textwidth}
    \includegraphics[width=\linewidth]{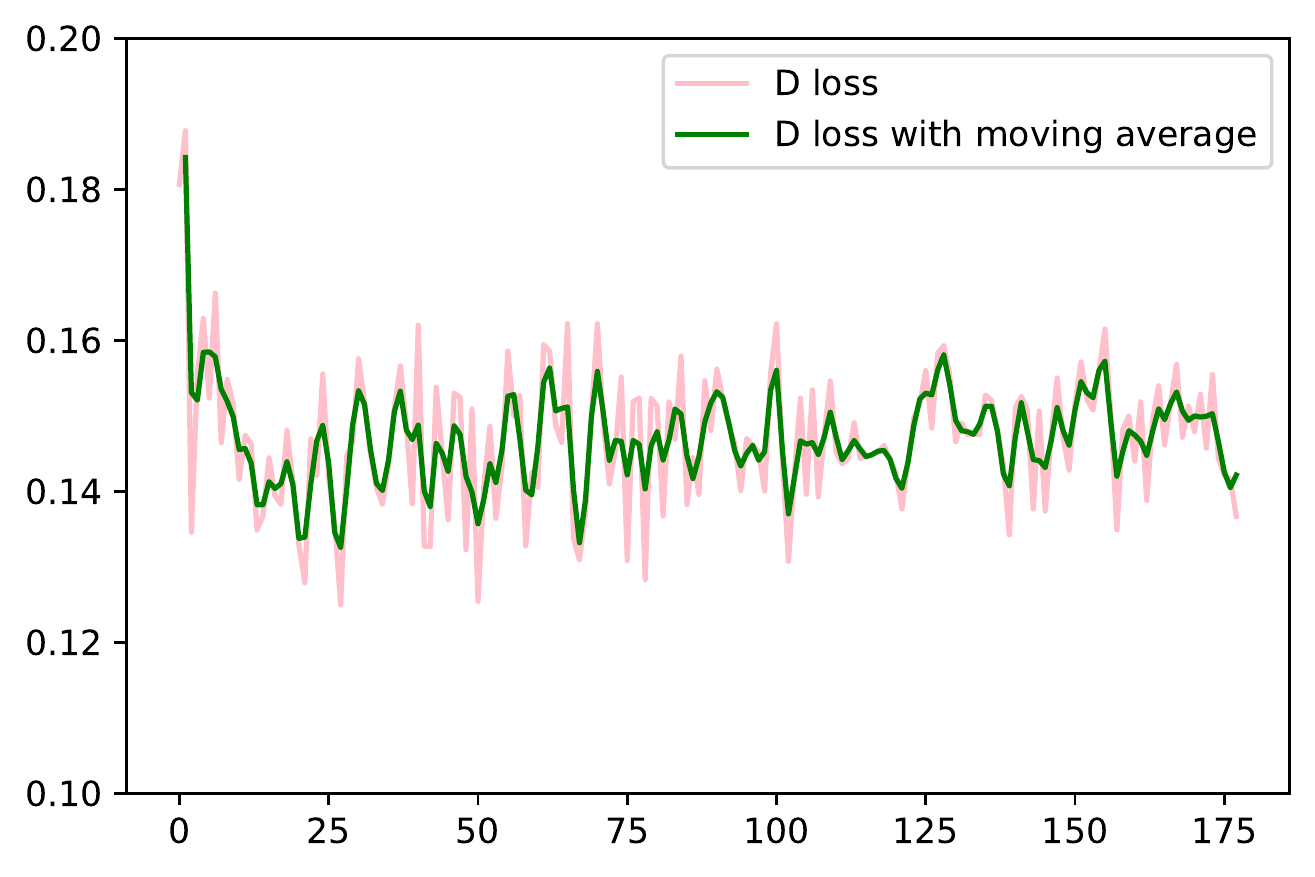}
\endminipage \hfill
\minipage{0.24\textwidth}
  \includegraphics[width=\linewidth]{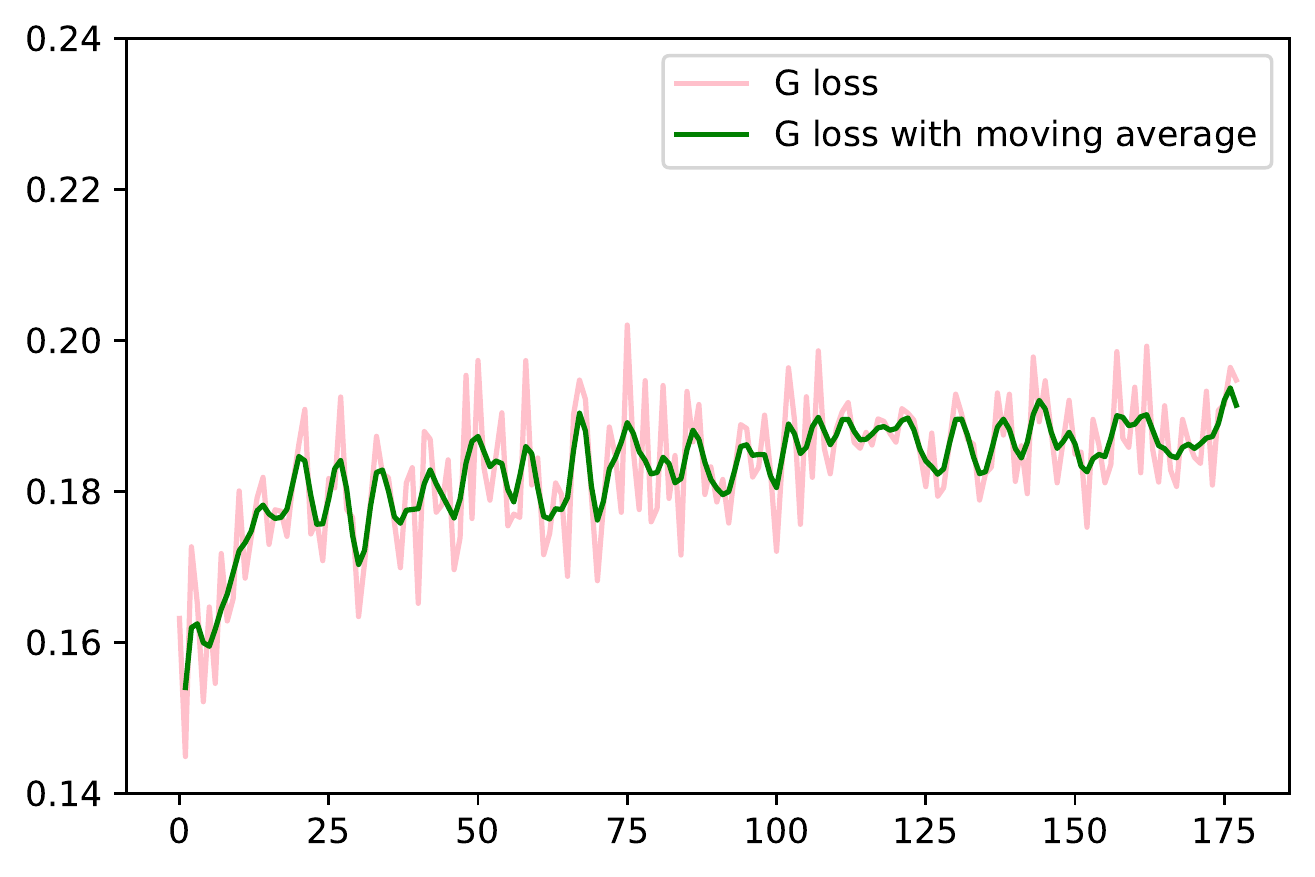}
\endminipage\hfill

\caption{ The discriminator (left) and generator (right) losses during training for CelebA. Each epoch is split into 25 averaged batch losses. The window size is 2 for the moving average.} 
\label{fig:gan_plots}
\end{figure}

\subsection{Applications of Clustering on Sensor Data}
In order to demonstrate the capabilities of our distributional clustering method on real data, we chose to experiment with activity classification using time-varying sensor data. Though many datasets are applicable, we chose to use WHARF~\cite{wharf}, MHEALTH~\cite{mhealth1, mhealth2}, and WISDM~\cite{wisdm} in our initial experiments. These datasets are collections of multimodal body sensor recordings as test subjects perform different activities of daily living (ADL), including but not limited to sitting, standing, eating, walking, and jogging.

The experimental setup is the same as in Section 5.1, where we compute the rand index and adjusted rand index scores on the test set in each experiment, averaged over 10 runs for each of the three methods. Results are given in Table~\ref{tab:appclust}. We note that only experiments using contrastive loss are reported here; though our distributional loss formula has the potential to be applied directly here, we leave this as future work.

\begin{figure}[t!]
\minipage{0.23\textwidth}
    \includegraphics[width=\linewidth]{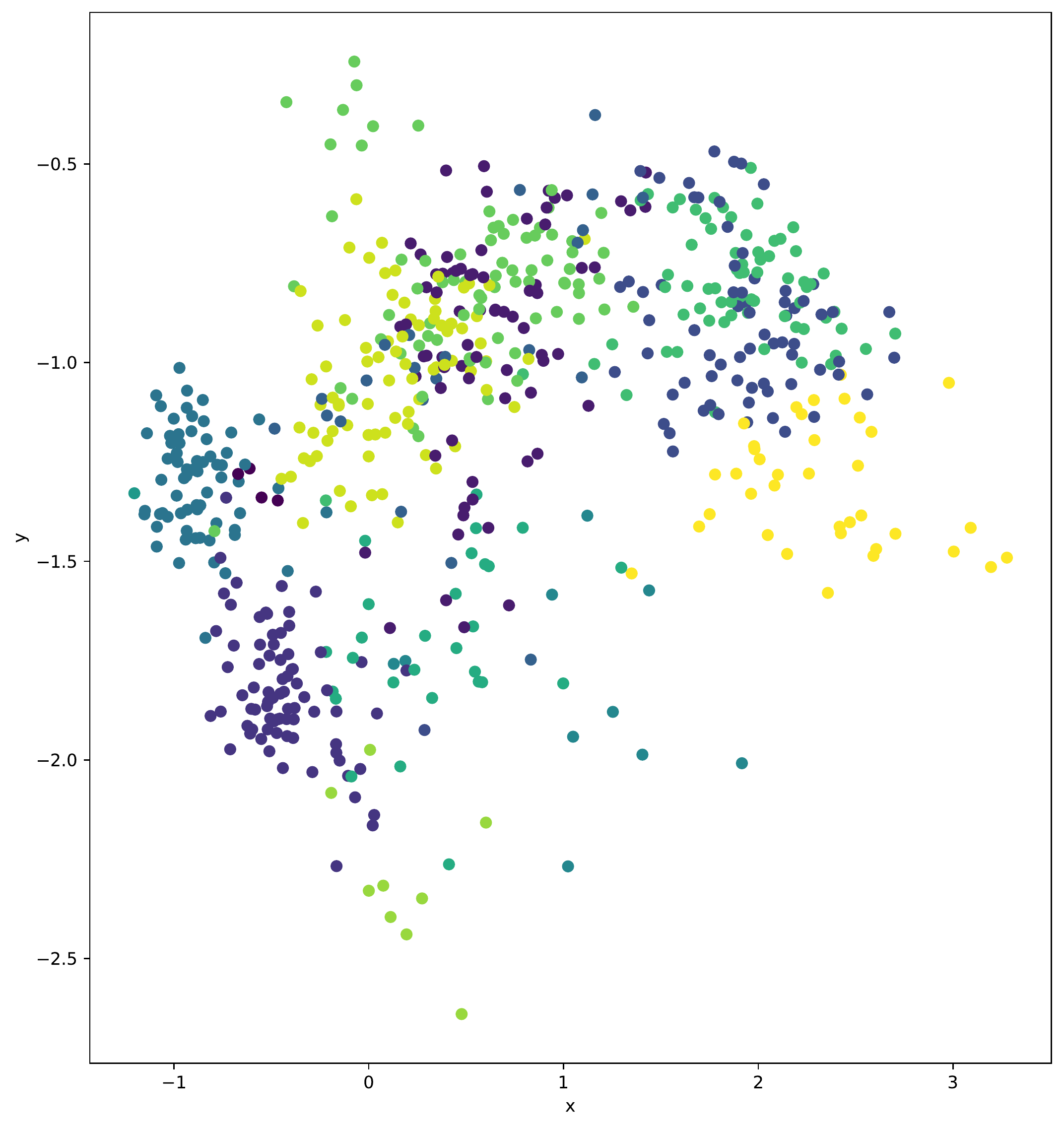}
\endminipage \hfill
\minipage{0.23\textwidth}
\includegraphics[width=\linewidth]{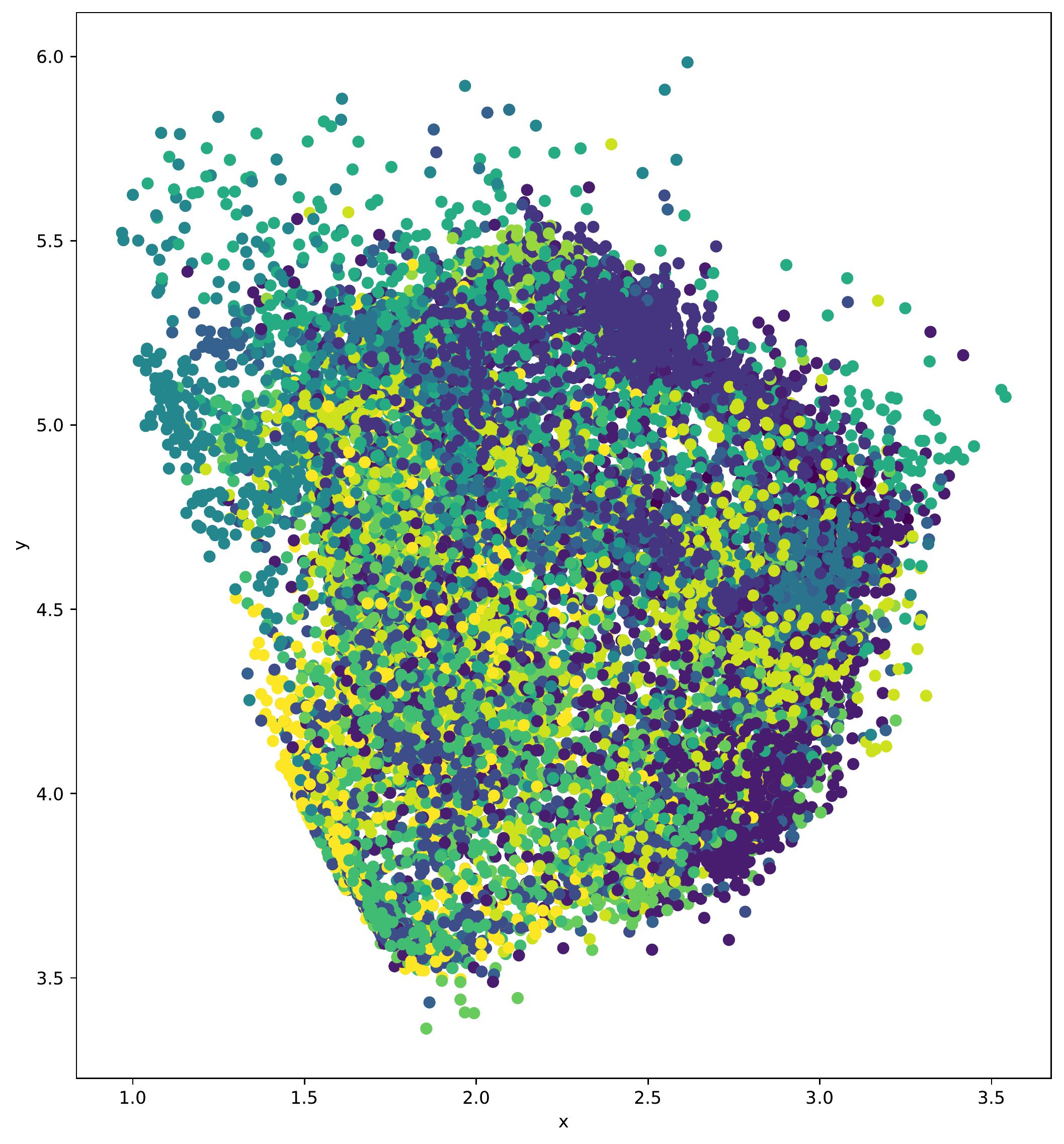}
\endminipage\hfill
\vskip\baselineskip
\caption{(Left) Embedding of the WHARF dataset learned by our method using contrastive loss with a moment-matching function. (Right) Embedding learned by the baseline deep learning approach.}
\label{fig:real-embed}
\end{figure}

As in Section 5.1, we visualize embeddings learned by our method and the baseline method, where we see that our learned embeddings capture the correct cluster structure, as pictured in Figure~\ref{fig:real-embed}.

\begin{table}[t]
\centering
\resizebox{1\columnwidth}{!}{%
\renewcommand{\arraystretch}{1.5}
\begin{tabular}{cccclclc}
\hline
\multirow{2}{*}{\textbf{Dataset}} &  &  & \multicolumn{2}{c}{\multirow{2}{*}{Baseline Method}} & \multicolumn{2}{c}{\multirow{2}{*}{Our Method}} & \multirow{2}{*}{\begin{tabular}[c]{@{}c@{}}David \& \\  Dhillon\end{tabular}} \\
 &  &  & \multicolumn{2}{c}{} & \multicolumn{2}{c}{} &  \\ \hline 
\hline
\multirow{4}{*}{WHARF} & \multirow{2}{*}{RI} & Mean & \multicolumn{2}{c}{0.832} & \multicolumn{2}{c}{\textbf{0.887}} & 0.876 \\
 &  & Std & \multicolumn{2}{c}{0.002} & \multicolumn{2}{c}{0.004} & 0.007 \\
 & \multirow{2}{*}{ARI} & Mean & \multicolumn{2}{c}{0.098} & \multicolumn{2}{c}{\textbf{0.364}} & 0.327 \\
 &  & Std & \multicolumn{2}{c}{0.006} & \multicolumn{2}{c}{0.022} & 0.026 \\ \hline
\multirow{4}{*}{MHEALTH} & \multirow{2}{*}{RI} & Mean & \multicolumn{2}{c}{0.849} & \multicolumn{2}{c}{\textbf{0.860}} & 0.664 \\
 &  & Std & \multicolumn{2}{c}{0.005} & \multicolumn{2}{c}{0.007} & 0.006 \\
 & \multirow{2}{*}{ARI} & Mean & \multicolumn{2}{c}{0.106} & \multicolumn{2}{c}{\textbf{0.149}} & 0.023 \\
 &  & Std & \multicolumn{2}{c}{0.008} & \multicolumn{2}{c}{0.018} & 0.001 \\ \hline
\multirow{4}{*}{WISDM} & \multirow{2}{*}{RI} & Mean & \multicolumn{2}{c}{0.894} & \multicolumn{2}{c}{\textbf{0.907}} & 0.900 \\
 &  & Std & \multicolumn{2}{c}{0.004} & \multicolumn{2}{c}{0.003} & 0.003 \\
 & \multirow{2}{*}{ARI} & Mean & \multicolumn{2}{c}{0.086} & \multicolumn{2}{c}{\textbf{0.127}} & 0.089 \\
 &  & Std & \multicolumn{2}{c}{0.005} & \multicolumn{2}{c}{0.014} & 0.009 \\ \hline
\end{tabular}
}
\caption{Rand index and adjusted rand index scores for different clustering experiments performed on real data, where the baseline method treats each training point independently.} 
\label{tab:appclust}
\end{table}

 \subsection{Additional K-nn Classification Details}
Here we provide more details regarding our K-nn experiments between deep Bregman and Euclidean cases. All factors in our experimental settings are created by very standard choices for a fair comparison. The batches are chosen randomly from the relevant dataset, and then the pairs are created within that batch at each iteration. We use a validation set ratio of $20 \%$. Once the training is complete, we obtain test embeddings and run the K-nn algorithm on these embeddings.

We choose $k$, the number of subnetworks, to be equal to the number of classes. 
Additionally, we run a small experiment over varying $k$ from 5 to 1000 and reported the results in Table~\ref{tab:varyingk}. The results indicate that performance improves to a point, and then the model starts to overfit. This suggests that an optimal $k$ can be found by adding it as a hyperparameter in the experiments. 
\begin{table}[t]
\centering
\resizebox{.95\columnwidth}{!}{%
\renewcommand{\arraystretch}{1.3}
\begin{tabular}{cccccccc}
\hlineB{1.75}
$k$  & 5  & 20 & 50 & 100 & 200 & 500 & 1000 \\ 
 \hline
acc & 71.9 & 77.8  & 79.4 & 80.0 & 77.4 & 74.1 & 70.8\\ \hlineB{1.75}
\end{tabular} 
}
\caption{ Accuracy on Cifar10 when varying the number of subnetworks ($k$). } 
\label{tab:varyingk}
\end{table}  

\end{appendices}

\end{document}